\documentclass{article}
\pdfoutput=1

     \usepackage[nonatbib,final]{neurips_2019}

\usepackage[utf8]{inputenc} 
\usepackage[T1]{fontenc}    
\usepackage{hyperref}       
\usepackage{url}            
\usepackage{booktabs}       
\usepackage{amsfonts}       
\usepackage{nicefrac}       
\usepackage{microtype}      

\usepackage{xcolor}
\usepackage{amsmath}
\usepackage{breqn}
\usepackage{float}
\usepackage{graphicx}
\usepackage{caption}
\usepackage{subcaption}
\usepackage[cal=cm]{mathalfa}

\usepackage[linesnumbered,ruled]{algorithm2e}
\usepackage[noend]{algpseudocode}
\usepackage[normalem]{ulem}

\usepackage{comment}
\usepackage{multirow}
\usepackage{amssymb}
\usepackage{amsfonts}
\usepackage{mathtools}
\DeclarePairedDelimiter\ceil{\lceil}{\rceil}
\DeclarePairedDelimiter\floor{\lfloor}{\rfloor}
\def\l1{$\ell_1$}

\makeatletter

\def\algbackskip{\hskip-\ALG@thistlm}
\makeatother

\graphicspath{ {fig/} }

\title{Exploring Bit-Slice Sparsity in Deep Neural Networks for Efficient ReRAM-Based Deployment}

\author{Jingyang Zhang\textsuperscript{\rm 1}, Huanrui Yang\textsuperscript{\rm 1}, Fan Chen\textsuperscript{\rm 1}, Yitu Wang\textsuperscript{\rm 2}, Hai Li\textsuperscript{\rm 1}\\ \textsuperscript{\rm 1}Department of Electrical and Computer Engineering, Duke University\\\textsuperscript{\rm 2}School of Microelectronics, Fudan University\\\{jingyang.zhang, huanrui.yang, fan.chen, hai.li\}@duke.edu, ytwang16@fudan.edu.cn}

\begin{document}

\maketitle

\begin{abstract}
Emerging resistive random-access memory (ReRAM) has recently been intensively investigated to accelerate the processing of deep neural networks (DNNs).
Due to the \textit{in-situ} computation capability, analog ReRAM crossbars yield significant throughput improvement and energy reduction compared to traditional digital methods.
However, the power hungry analog-to-digital converters (ADCs) prevent the practical deployment of ReRAM-based DNN accelerators on end devices with limited chip area and power budget.
We observe that due to the limited bit-density of ReRAM cells, DNN weights are bit sliced and correspondingly stored on multiple ReRAM bitlines.
The accumulated current on bitlines resulted by weights directly dictates the overhead of ADCs.
As such, bitwise weight sparsity rather than the sparsity of the full weight, is desirable for efficient ReRAM deployment.
In this work, we propose \textit{bit-slice \l1}, the first algorithm to induce bit-slice sparsity during the training of dynamic fixed-point DNNs. 
Experiment results show that our approach achieves 2$\times$ sparsity improvement compared to previous algorithms. 
The resulting sparsity allows the ADC resolution to be reduced to 1-bit of the most significant bit-slice and down to 3-bit for the others bits, which significantly speeds up processing and reduces power and area overhead. 
\end{abstract}

\vspace{-10pt}
\section{Introduction}
\label{sec:intro}
\vspace{-5pt}
Although the promising performance of Deep neural network (DNN) models have been demonstrated in various real-world tasks~\cite{simonyan2014very,makantasis2015deep}, the intensive computing and memory requirements of DNN processing make its deployment extremely difficult , especially on end devices with limited resources and rigid power budget~\cite{han2015learning,wen2016learning}. 
The challenges of efficient deployment of large DNN models have motivated researches on model compression, including
pruning~\cite{han2015learning,wen2016learning} and quantization~\cite{gysel2016ristretto,polino2018model}.
Coupled with algorithm development, customized CMOS DNN accelerators are extensively investigated to take full advantage from model compression algorithms. 
For example, ESE~\cite{ese2017han} is optimized to achieve high computation efficiency on element-wise sparse DNNs, while DNPU~\cite{dnpu2017shin} supports low-precision, dynamic fixed-point operations.
However, these digital approaches typically require that most of the network weights to be stored off-chip, resulting in a large performance penalty for memory access.

In the meantime, the emerging resistive random-access memory (ReRAM) provides a novel mixed-signal design paradigm.
In general, DNN weights are encoded as the ReRAM cell conductance,  while the core computing pattern in DNN processing, i.e., massive matrix-vector multiplications, can be executed \textit{in-situ} in one computing-in-memory (CIM) cycle without moving data back and forth in the memory hierarchy.
Indeed, prior ReRAM-based accelerators have demonstrated two orders of magnitude advantages in energy, performance and chip footprint, over their digital counterparts~\cite{shafiee2016isaac, song2017pipelayer}.
However, the conversion between digital and analog domains, especially the analog-to-digital converters (ADCs), limit the effectiveness of these CIM designs to a certain extent because they normally account for $>60\%$ power and $>30\%$ area overhead~\cite{shafiee2016isaac}.

We observe that each operand (i.e. weight) is bit-sliced across multiple ReRAM bitlines (located in the same row) due to the limited cell bit density. 
The accumulated currents on bitlines dictate ADC's bit-resolution, which in turn determines the size and power consumption of the ADCs, as the ADC overhead generally increases exponentially with its resolution~\cite{shafiee2016isaac}. 
Higher sparsity in each bit-slice is desired to reduce the accumulated currents.
Based on this observation, we propose \emph{bit-slice \l1}, a novel sparsity regularization that applies \l1 penalization to all bit-slices of each fixed-point weight elements during training to induce bit-slice sparsity.
Unlike previous weight-grade sparsity methods, such finer-grained sparsity distribution achieves balanced sparsity when mapped to practical ReRAM crossbars, resulting in efficient deployment and significantly reduced ADC overhead.
Existing work explored bit-partition~\cite{Ghodrati2019mixedsignal} and dynamic bit-level fusion/decomposition~\cite{Sharma2018bitfusion} in efficient DNN accelerator designs, but none of these works considered the sparsity within each bit-slice.
Therefor, our work on bit-slice sparsity provides new opportunities to effectively exploit sparsity in sparse accelerators, as initially demonstrated in~\cite{Yang2019sre}.

We apply \emph{bit-slice \l1} regularization to the training process of a 8-bit dynamic fixed-point DNN to show its effectiveness.
We assume each slice contains two bits because 2 bits/cell is the most common muti-level cell type in current ReRAM technology.
Please note that as technology advances, our approach can be easily extended to support more bits per slice.
Experiment results shows that \emph{bit-slice \l1} achieves $2\times$ sparsity improvement compared to previous full-number pruning algorithms. 
The resulting sparsity can significantly speeds up processing and reduces power and area overhead in ReRAM deployment. 
To the best of our knowledge, this is the first algorithm specifically designed to train DNN models that are friendly for the bit sliced deployment on ReRAM crossbars.

\vspace{-10pt}

\section{Proposed method}
\label{sec:method}
\vspace{-5pt}
In this section, we first describe the procedure for training a DNN with dynamic fixed-point quantization, which fits the requirement for ReRAM deployment. Then we introduce our bit-slice \l1 regularizer, which aims at providing bit-slice sparsity for the ReRAM instead of the sparsity on the full-number weights. Finally, we present the whole training routine of our method as we apply the proposed regularizer to the dynamic fixed-point training process. The training routine and the proposed regularizer is demonstrated in Figure~\ref{fig:routine}.

\vspace{-5pt}
\subsection{Dynamic fixed-point quantization}
\vspace{-5pt}
As observed by Polino et al.~\cite{polino2018model}, the weight of different layers may have various dynamic ranges. Keeping the dynamic range of each layer is important for maintaining the performance of the model, especially after low-precision quantization~\cite{polino2018model}. Therefore for each layer, we need to first compute its dynamic range and scale the weight to the range of $[0,1]$ before applying quantization. Since state-of-the-art ReRAM based accelerators often map negative and positive weight elements to separated crossbars~\cite{song2017pipelayer}, here we ignore the sign of weight elements and only focus on quantizing their absolute values. The dynamic range of a layer $W_l$ is defined as:
\begin{equation}
    S(W_l)=\ceil{\log_2(\max_{w_l^i\in W_l}(|w_l^i|))},
\end{equation}
where $i$ is the index of each weight element in layer $l$. 

Then we apply uniform quantization to the scaled weight. Considering the scaling factor $2^{-S(W_l)}$ applied on the weight elements, the quantization step size of a $n$-bit quantization would be $Q_{step}=2^{S(W_l)-n}$, and, the weight element $w_l^i$ will be quantized to:
\begin{equation}
\label{equ:quant}
    B(w_l^i) = \floor{\frac{w_l^i}{Q_{step}}}.
\end{equation}
This mapping will guarantee all $B(w_l^i)$ are within the range of $[0,2^n-1]$. The $n$-bit binary representation of $B(w_l^i)$ will be stored in the ReRAM for computation. The dynamic range of the original weight can be recovered as $Q(w_l^i)=B(w_l^i)\cdot Q_{step}$, which can be easily implemented with a shifting operation after each layer's computation on the ReRAM crossbar. The quantization precision is set to 8 bits in this work, which is efficient in hardware deployment without significant accuracy loss.

\vspace{-5pt}
\subsection{Bit-slice $\ell_1$}
\vspace{-5pt}
After quantization, the quantized weight $B(w_l^i)$ can be represented in binary form as $B(w_l^i)=\sum_{j=0}^7b_j\cdot 2^j$. Then for ReRAM mapping, $B(w_l^i)$ will be sliced into four 2-bit slices, i.e. $\{b_7,b_6\}$, $\{b_5,b_4\}$, $\{b_3,b_2\}$, and $\{b_1,b_0\}$ where $b_7$ is the MSB and $b_0$ is the LSB, and be mapped onto 4 separated crossbars. Here we propose the \emph{bit-slice \l1} regularizer, which can apply \l1 regularization to all the bit-slices simultaneously in order to reach a sparse mapping on all ReRAM crossbars. Formally, the bit-slice process can be represented as $B(w_l^i)=\sum_{k=0}^{3}\hat{B}_l^{i,k}\cdot2^{2k}$,
where $\hat{B}_l^{i,k}$ will be an integer within the range of $[0, 3]$. The bit-sliced \l1 of weight $W_l$ is therefore defined as:
\begin{equation}
    B\ell_1(W_l) := \sum_{i,k} \hat{B}_l^{i,k}.
\end{equation}
Note that the $B\ell_1$ regularizer takes the full weight $W_l$ as input for training. This property enables the regularizer to smoothly fit into the training routine of a dynamic fixed-point DNN.

\subsection{Training routine}
The proposed $B\ell_1$ regularizer enables us to achieve bit-slice sparsity by training from scratch. Yet we find it would be more efficient in reaching higher sparsity by starting from a pretrained, element-wise sparse model, such as a model trained with the \l1 regularizer. 

\begin{figure}[!t]
\centering
\includegraphics[width=0.7\linewidth]{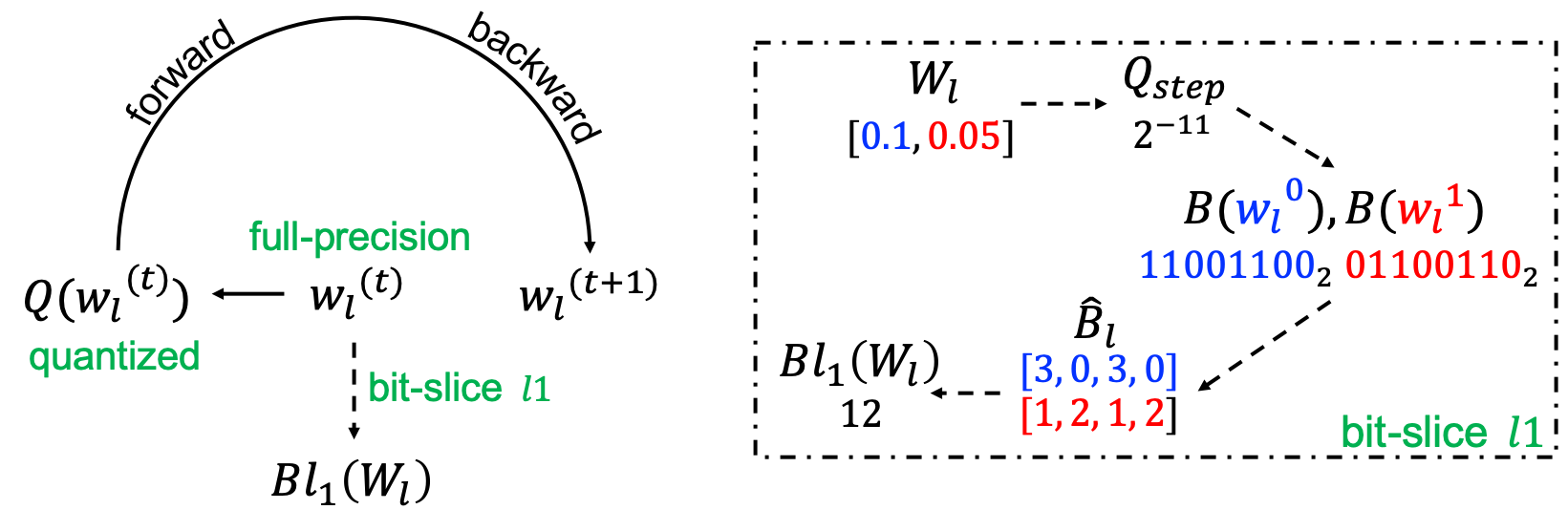}
\caption{Illustration of the training routine and our bit-slice $\ell_1$.}
\label{fig:routine}
\vspace{-10pt}
\end{figure}

We follow the training procedure proposed in~\cite{gysel2016ristretto} to train the dynamic fixed-point network. Specifically, we keep full precision weights during the training. As shown in Figure~\ref{fig:routine}, for each step, we first quantize $w_l^i$ to $B(w_l^i)$ as described in Equation~(\ref{equ:quant}), then the $w_l^i$ is replaced with the recovered quantized weight $Q(w_l^i)$. We use $Q(w_l^i)$ to do the forward pass, compute the cross entropy loss $\mathcal{L}_{CE}$ and the penalty of the $B\ell_1$ regularizer. The gradient is then accumulated to $Q(w_l^i)$ with full precision, which will be used as the new $w_l^i$ for the next step. The update rule for each training step can be formally formulated as (index $i$ is omitted here for clarity):
\begin{equation}
    q^{(t)} = Q(w_l^{(t)}), \ \ w_l^{(t+1)} = q^{(t)}-lr\times(\nabla_q \mathcal{L}_{CE}(q^{(t)})+\alpha \nabla_q B\ell_1(q^{(t)}))
\end{equation}

\vspace{-10pt}
\section{Experiment results}
\label{sec:exp}

We test the proposed bit-slice \l1 on the MNIST benchmark \cite{lecun1998gradient} with a toy model consisting of two linear layers, and on the CIFAR-10 dataset \cite{krizhevsky2009learning} with VGG-11 \cite{simonyan2014very} and ResNet-20 \cite{he2016deep}. All the models are implemented and trained in the deep learning framework PyTorch\footnote{Codes for the experiments available at \url{https://github.com/zjysteven/bitslice_sparsity}}.

\begin{table}[tb]
    \caption{Results on MNIST}
    \label{tab:mnist}
    \centering
    \begin{tabular}{ccccccc}
    \toprule
        & &\multicolumn{5}{c}{Ratio of non-zero wights}         \\
                \cmidrule(r){3-7}
    Method &Accuracy  &$\hat{B^3}$ &$\hat{B^2}$ &$\hat{B^1}$ &$\hat{B^0}$ &Average\\
    \midrule
    Pruned &97.99\% &1.08\% &5.87\%  &8.42\% &17.42\% &8.20$\pm$5.94\%\\
    \midrule
    \l1 &97.99\% &1.19\% &5.21\% &7.01\% &11.36\% &6.19$\pm$3.65\%   \\
    B\l1 &97.67\% &\textbf{0.84\%}  &\textbf{4.02\%}  &\textbf{4.27\%} &\textbf{9.58\%} &\textbf{4.68$\pm$3.14\%}    \\
    \bottomrule
    \end{tabular}
    \vspace{-10pt}
\end{table}

\begin{table}[tb]
    \caption{Results on CIFAR-10}
    \label{tab:cifar}
    \centering
    \begin{tabular}{cccccccc}
    \toprule
     & & &\multicolumn{5}{c}{Ratio of non-zero wights}         \\
                \cmidrule(r){4-8}
    Model &Method &Accuracy  &$\hat{B^3}$ &$\hat{B^2}$ &$\hat{B^1}$ &$\hat{B^0}$ &Average\\
    \midrule
    \multirow{3}{*}{VGG-11} &Pruned &88.93\% &0.86\% &28.30\%  &34.14\% &33.39\% &24.17$\pm$13.65\%\\ \cmidrule(r){2-8}
    &\l1 &\textbf{89.39\%} &0.39\% &9.37\% &18.43\% &22.19\%  &12.59$\pm$8.45\%  \\
    &B\l1 &89.33\% &\textbf{0.21\%}  &\textbf{3.57\%}  &\textbf{7.09\%} &\textbf{10.71\%}    &\textbf{5.40$\pm$3.92\%} \\
    \midrule\midrule
    \multirow{3}{*}{ResNet-20} &Pruned &89.22\% &1.10\% &8.07\%  &21.92\% &43.96\% &18.76$\pm$16.36\%\\ \cmidrule(r){2-8}
    &\l1 &\textbf{90.62\%} &0.44\% &4.71\% &14.37\% &33.16\% &13.17$\pm$12.60\%   \\
    &B\l1 &89.66\% &\textbf{0.31\%}  &\textbf{3.34\%}  &\textbf{11.99\%} &\textbf{31.39\%}    &\textbf{11.76$\pm$12.12\%} \\
    \bottomrule
    \end{tabular}
    \vspace{-10pt}
\end{table}

Table \ref{tab:mnist} and Table \ref{tab:cifar} summarize the performance of the proposed bit-slice \l1 on MNIST and CIFAR-10, respectively. Note that in these two tables and the rest of this section, $\hat{B}^3$, $\hat{B}^2$, $\hat{B}^1$, and $\hat{B}^0$ represent the 4 slices of the bit-slice weights, from the most significant to the least significant respectively. The sparsity is computed across the whole model. We take normal \l1 regularization as a baseline, which is applied to the full weight. Without regularization, $\hat{B}^3$ can be pruned to about 1\% non-zero weights, while $\hat{B}^2$, $\hat{B}^1$, and especially $\hat{B}^0$ still have a large amount of non-zero elements, resulting in significant imbalance between the sparsity of bit-slice weights. The element-wise sparsity induced by normal \l1 regularization is able to improve bit-slice sparsity; while our bit-slice \l1 achieves higher sparsity in all test cases. Bit-slice \l1 also mitigates the unbalanced sparsity as shown by the lower standard variance compared to normal \l1's results. These results support that the proposed bit-slice \l1 fits better for ReRAM, to which bit-slice sparsity is of great importance.

\begin{figure}[!t]
\centering
\includegraphics[width=0.9\linewidth]{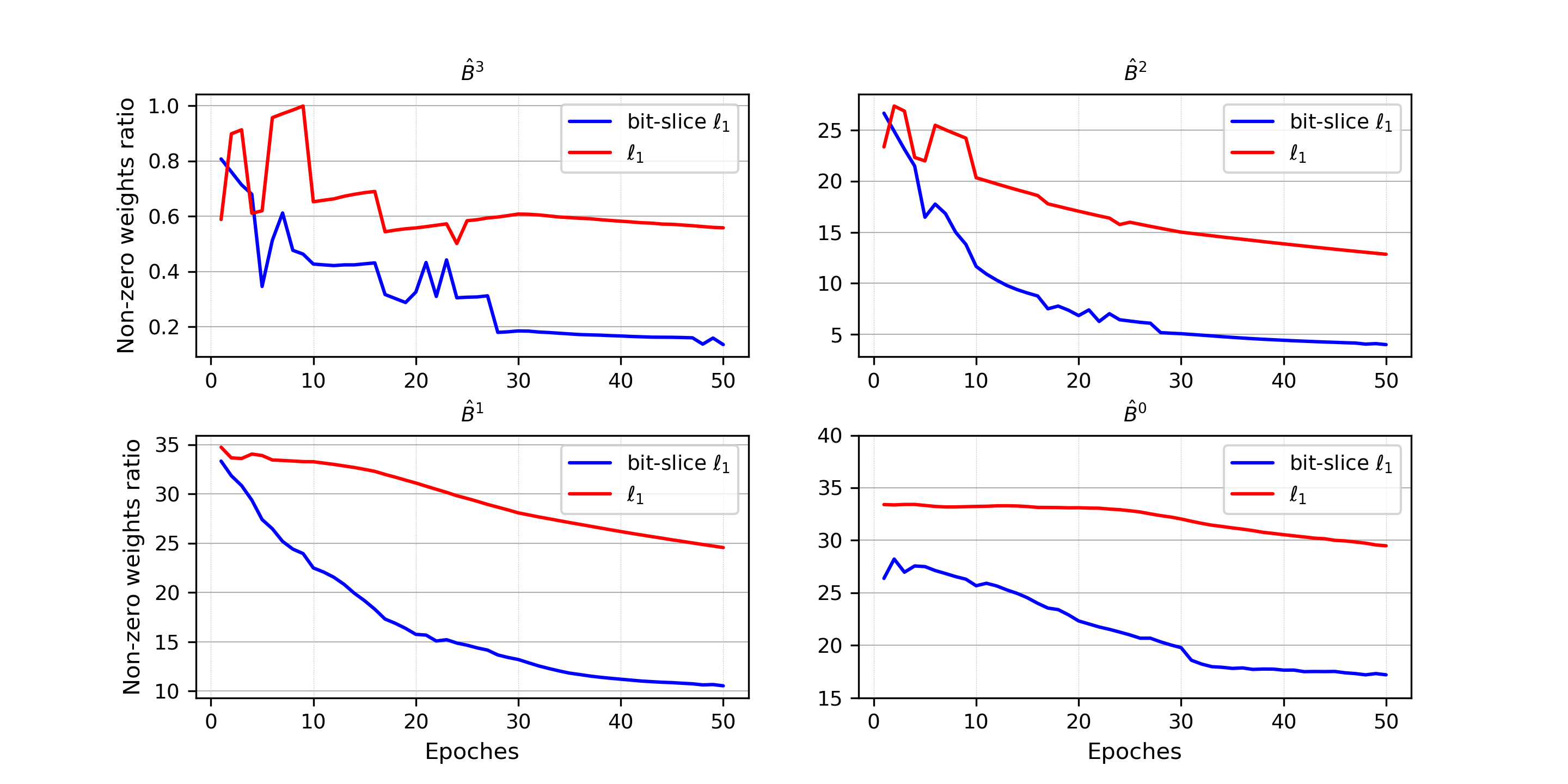}
\caption{Bit-slice sparsity of VGG-11 on CIFAR-10 during training.}
\label{fig:curve}
\vspace{-10pt}
\end{figure}

Figure~\ref{fig:curve} compares the percentage of non-zero bit-slice elements during the training with original \l1 and the proposed bit-slice \l1. It can be clearly observed that bit-slice \l1 reduces the number of non-zero bit-slices faster than normal \l1 regularization from the very beginning, which again proves that bit-slice \l1 is a better option for regularizing bit-slice weights.

\begin{table}[tb]
    \caption{ADC Overhead Saving with Bit-Slice Sparsity}
    \label{tab:ADC}
    \centering
    \begin{tabular}{c|c|cccc}
    \toprule
     & w/o Bit-Slice Sparsity & \multicolumn{4}{c}{w/ Bit-Slice Sparsity}         \\
                \cmidrule(r){2-6}
     & Resolution & Resolution & Energy Saving & Speedup & Area Saving \\
    \midrule
    XB$_{3}$ & 8 bit & 1 bit & 28.4$\times$ & 8$\times$ & 2$\times$ \\
    \midrule
    XB$_{2,1,0}$ & 8 bit & 3 bit & 14.2$\times$ & 2.67$\times$ & 2$\times$ \\
    \bottomrule
    \end{tabular}
    \vspace{-10pt}
\end{table}

In simulation, we map the achieved 8-bit weights with bit-slice sparsity onto 4 groups of $128\times128$ ReRAM crossbars (XBs), with each group storing 2 bits of the 8-bit weights. XB$_{3,2,1,0}$ store the 2-bit slices from the MSB to the LSB respectively. According to the sparsity level achieved, we can apply ADCs with different resolutions to the 4 groups of XBs. As illustrated in Table~\ref{tab:ADC}, ISAAC~\cite{shafiee2016isaac} needs to use 8-bit ADCs to store the weights without bit-slice sparsity even after ADC optimization. However, with the bit-slice sparsity achieved by bit-slice \l1, we can use 1-bit and 3-bit ADCs instead. According to ~\cite{ADC}, the power of ADC is approximately proportional to $2^N/(N+1)$ and the sensing time of ADC is directly proportional to N. Here, N denotes the resolution of ADC. Therefore, with bit-slice sparsity, the 1-bit ADC of XB$_{0}$ can achieve $28.4\times$ energy saving and $8\times$ sensing time speedup. Meanwhile, the 3-bit ADC can achieve $14.2\times$ energy saving and $2.67\times$ sensing time speedup. From the area perspective, the area of a 6-bit ADC is approximately the half of an 8-bit ADC but the area varies little when the resolution is lower than 6. Thus, with bit-slice sparsity, the ADC can achieve $2\times$ area saving.

\vspace{-5pt}
\section{Conclusion}
\vspace{-5pt}
In conclusion, we proposed bit-slice $\ell_1$ regularizer, the first algorithm specifically designed to train DNN models that are friendly for the bit sliced deployment on ReRAM crossbars. The proposed method can induce higher and more balanced sparsity levels among bit-slices of DNN weights comparing to traditional element-wise sparsity inducing training methods. The achieved bit-slice sparsity will enable significant reduction on the ADC energy and area overhead, and will further improve inference speed for ReRAM deployment.

\vspace{-10pt}
\subsubsection*{Acknowledgement}
\vspace{-5pt}
This work was supported in part by NSF CNS-1822085 and NSF CSR-1717885.
\vspace{-5pt}


\small

\setlength{\parskip}{0.0pt}
\setlength{\itemsep}{0.0pt}
\bibliographystyle{unsrt}

\end{document}